\documentclass[letterpaper]{article} 
\usepackage{aaai2027}  
\usepackage[hyphens]{url}  
\usepackage{graphicx} 
\usepackage{natbib}  
\usepackage{caption} 
\usepackage{algorithm}
\usepackage{algpseudocode}
\algrenewcommand\algorithmicrequire{\textbf{Input:}}
\algrenewcommand\algorithmicensure{\textbf{Output:}}
\usepackage{amsmath}
\usepackage{amssymb}
\usepackage{threeparttable}
\newcommand{\cmark}{\checkmark}

\usepackage{newfloat}
\usepackage{listings}
\usepackage{booktabs}
\usepackage{multirow}
\usepackage{makecell}
\usepackage{adjustbox}
\usepackage[table]{xcolor}
\usepackage{subcaption}
\usepackage{caption}
\DeclareCaptionStyle{ruled}{labelfont=normalfont,labelsep=colon,strut=off} 
\floatstyle{ruled}
\newfloat{listing}{tb}{lst}{}
\floatname{listing}{Listing}
\usepackage{xcolor}
\usepackage[most]{tcolorbox}

\definecolor{gaincolor}{RGB}{0,135,110}
\definecolor{losscolor}{RGB}{220,35,35}

\newcommand{\gain}[1]{%
    \,\textcolor{gaincolor}{\ensuremath{\uparrow}\,#1}%
}
\newcommand{\loss}[1]{%
    \,\textcolor{losscolor}{\ensuremath{\downarrow}\,#1}%
}
\title{CASE: Confusion-Aware Skill Evolution for Frozen Multimodal Emotion Models}
\author{
    Siyu Wu\textsuperscript{1},
    Zeyu Liang\textsuperscript{1},
    Linghao Wang\textsuperscript{1},
    Qiyun Zheng\textsuperscript{1},
    Yi Yao\textsuperscript{2},
    Ling Lo\textsuperscript{3},
    Hongxia Xie\textsuperscript{1}
}

\affiliations{
    \textsuperscript{1}Jilin University\\
    \textsuperscript{2}National Chiao Tung University\\
    \textsuperscript{3}National Tsing Hua University
}

\begin{document}

\maketitle
\begin{abstract}

Multimodal large language models (MLLMs) have substantially advanced Multimodal Emotion Recognition (MER) by enabling fine-grained emotion understanding through reasoning over multimodal evidence. However, they still exhibit systematic confusion between semantically adjacent emotions with highly overlapping multimodal cues. Existing approaches typically address these errors through global parameter adaptation, making confusion correction costly and inflexible. We instead observe that emotion confusion is highly localized and therefore better addressed through reusable pair-specific expertise. Building on this insight, we propose Confusion-Aware Skill Evolution (CASE), the first agent-skill framework for MER. CASE maintains a sparse, self-evolving library of pair-specific EmotionSkills that externalize affective expertise from model parameters, enabling the learned expertise to be continuously refined and readily transferred across diverse frozen MLLM backbones. During inference, CASE dynamically retrieves the appropriate EmotionSkill to reinforce boundary-aware reasoning while keeping the MLLM backbone frozen.
Extensive experiments on CA-MER, DFEW, and EMER demonstrate the effectiveness and transferability of CASE across both general-purpose and emotion-specialized MLLMs. Notably, CASE improves the average accuracy of Qwen3-VL-8B on the CA-MER benchmark from 46.80\% to 59.13\%, while achieving consistent gains across MLLMs. 
\end{abstract}

\section{Introduction}
\label{sec:introduction}


Recent advances in multimodal large language models (MLLMs) have shown impressive performance on video-based Multimodal Emotion Recognition (MER). Beyond predicting emotion labels directly, MLLMs can integrate visual, temporal, and contextual evidence to provide interpretable explanations for their decisions, making them a promising foundation for grounded affective computing~\citep{lian2023emer,cheng2024emotionllama}. 

Despite this progress, fine-grained emotion recognition remains a fundamental challenge. Although current MLLMs often identify the relevant affective evidence, they still struggle to distinguish semantically similar emotions, such as \textit{worry} versus \textit{sadness} or \textit{fear} versus \textit{surprise}~\citep{lian2025affectgpt,hu2025beyondemotion}. This limitation suggests that the bottleneck of modern MER has shifted from coarse affective perception to fine-grained boundary discrimination.

\begin{figure}[t]
    \centering
    \IfFileExists{Figures/CASE_fig1_vertical_2.png}{%
        \includegraphics[width=\columnwidth]{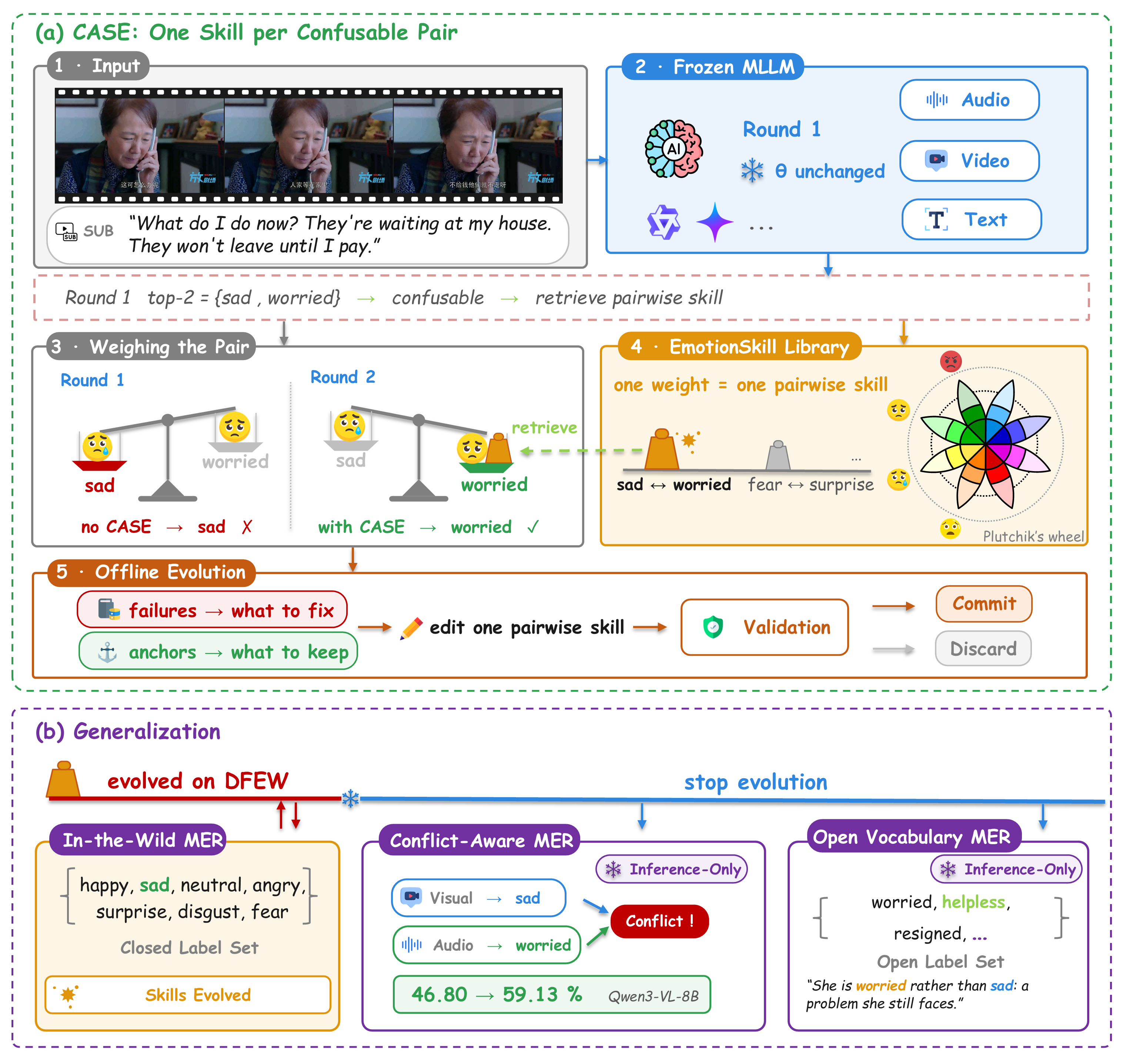}%
    }{%
        \includegraphics[width=0.8\columnwidth]{../Figures/CASE_fig1_vertical_2.png}%
    }
    \caption{Overview and cross-setting generalization of CASE.
        (a) CASE resolves Top-2 emotion ambiguity using a pair-specific skill.
        (b) Skills evolved only on closed-set in-the-wild DFEW are directly
transferred to audio-visual conflict-aware CA-MER and open-vocabulary
EMER, without target-dataset evolution or backbone updates.
    }
    \label{fig:case-teaser}
\end{figure}


More importantly, these misclassifications are not random prediction failures but exhibit a systematic structure. In psychological theories, emotions are characterized as a continuous affective space rather than isolated categories, where neighboring emotions naturally share semantic and behavioral characteristics~\citep{barrett2025constructed,plutchik2001nature}. The psychological structure is also reflected in MER, where prediction errors consistently concentrate on neighboring emotion pairs rather than being uniformly distributed across the label space. The observed structured confusion pattern suggests that improving fine-grained MER requires more than stronger general reasoning; it calls for mechanisms that explicitly reinforce discriminative boundaries between systematically confused emotion pairs. Existing approaches, however, primarily rely on global parameter adaptation or task-specific fine-tuning, implicitly treating all emotion categories equally. As a result, they overlook the localized nature of emotion confusion and lack an effective mechanism for accumulating and reusing pair-specific affective expertise.



Motivated by the systematic nature of emotion confusion, we propose \textbf{Confusion-Aware Skill Evolution (CASE)}, a general agent-skill framework that models fine-grained emotion recognition as a collection of pair-specific discrimination problems rather than a globally shared reasoning task. As shown in Figure~\ref{fig:case-teaser}, CASE maintains a sparse, self-evolving library of EmotionSkills. Each EmotionSkill encapsulates the discriminative knowledge required to resolve one systematically confused adjacent emotion pair. Specifically, CASE localizes ambiguous predictions to candidate confusion pairs, evolves confusion-specific skills based on recoverable errors, and selectively expands the skill library only when recurrent confusion patterns emerge. During inference, the most relevant EmotionSkill is dynamically retrieved to reinforce boundary-aware reasoning while keeping the MLLM backbone frozen. 
In this way, CASE externalizes affective expertise into modular knowledge that evolves independently of model parameters and can be readily reused across diverse frozen MLLM backbones.


More broadly, CASE shifts affective recognition from fixed label classification toward reusable knowledge-based reasoning, enabling frozen MLLMs to flexibly support evolving emotion taxonomies without continual model adaptation. Our contributions can be summarized as:
\begin{itemize}
    \item We introduce CASE, which reframes MER from global multiclass classification toward selective discrimination on ambiguous fine-grained emotion boundaries.

    \item We develop a recoverability-aware skill evolution mechanism that learns from boundary-specific affective confusion. By contrasting recoverable failures with valid anchors and verifying candidate skill updates, CASE enables reliable, sparse skill evolution atop a frozen multimodal backbone.

    \item We evaluate CASE across closed-set in-the-wild recognition
(DFEW), audio-visual conflict-aware recognition (CA-MER), and
open-vocabulary emotion reasoning (EMER). 
Skills evolved only on
DFEW directly improve CA-MER and EMER without target-dataset
evolution or backbone updates, demonstrating transfer across tasks,
label spaces, and MLLMs.
\end{itemize}

\section{Related Work}
\label{sec:related_work}


\subsection{Multimodal Emotion Recognition with MLLMs}
\label{sec:related_mer}

Recent MLLM-based emotion research has expanded affective label spaces and multimodal evidence modeling~\citep{guo2025stimuvar,hu2025beyondemotion}. 
Emotion-specific architectures and instruction-tuning strategies further improve general emotion understanding~\citep{cheng2024emotionllama,lian2025affectgpt,yang2025omniemotion,li2025emoverse}, open-vocabulary formulations relax predefined emotion taxonomies~\citep{lian2025ovmer}, and retrieval-augmented methods introduce external context for compound-emotion reasoning~\citep{wen2025raer}.
Conflict-aware methods additionally mitigate incongruent evidence through adaptive fusion or modality selection~\citep{wang2023crossattention,yu2026fuse}.
Despite these advances, existing methods primarily optimize one-shot prediction and rarely address recurrent errors between confusable emotion categories~\citep{han2025camer,hu2025beyondemotion}.
In this paper, CASE instead treats such errors as local decision-boundary failures. It uses pair-specific external knowledge to refine these boundaries while keeping the underlying MLLM frozen.

\subsection{Non-Parametric Learning through Memory and Skills}
\label{sec:related_memory_skills}

Non-parametric learning methods differ in how they represent and reuse experience. 
Memory-based approaches retrieve previous examples or feedback~\citep{madaan2022memprompt}, whereas reflection-based agents retain verbal summaries of past interactions~\citep{shinn2023reflexion,park2023generative}. 
Iterative self-feedback can refine individual outputs without additional training~\citep{madaan2023selfrefine}, while prompt evolution adapts reusable task instructions through performance-guided search~\citep{fernando2024promptbreeder}.
More recent methods distill experience into reusable instructions, workflows, toolboxes, or executable skills~\citep{wang2024voyager,zhao2024expel,fu2024autoguide,wang2024trove,wang2025awm}, with persistent skill libraries supporting continual refinement and verifier-guided evolution~\citep{wang2026skillx,zhou2026mementoskills,zhang2026coevoskills,suzgun2026dynamic}.
CASE adapts this paradigm to multimodal emotion recognition. Each EmotionSkill captures discriminative cues for one confusable emotion pair, is selectively activated by the model's top-2 predictions, and is retained only when it improves the corresponding local decision boundary.

\section{Method}
\label{sec:method}
\subsection{Problem Formulation}
\label{sec:problem_overview}

Let $x=(a,v,c)\in\mathcal{X}$ denote a multimodal emotion sample,
where $a$ is an audio track, $v$ is a video, and $c$ is optional
linguistic context such as subtitles. Given an emotion label space
$\mathcal{Y}=\{y_1,\ldots,y_K\}$, the task is to predict the
ground-truth emotion $y^*\in\mathcal{Y}$ for each sample.
We treat the recognizer as a frozen prompt-conditioned MLLM
interface $f_\theta:\mathcal{X}\times\mathcal{P}\rightarrow\mathcal{O}$,
where $\mathcal{P}$ is the prompt space and $\mathcal{O}$ is the
structured output space whose schema is determined by the prompt.
Every emotion-valued field of an output takes values in $\mathcal{Y}$.

The errors addressed by CASE are not caused by a complete
lack of emotion knowledge. Instead, the MLLM often
identifies two plausible emotions but fails to apply the appropriate
criterion for distinguishing them. In such cases, reconsidering all $K$ emotion
labels introduces unnecessary alternatives rather than resolving ambiguous emotion boundaries.

CASE therefore decomposes emotion recognition into local
discrimination over candidate confusion pairs, guiding frozen MLLMs to distinguish the two emotions through pair-specific EmotionSkills.
These skills are stored in a sparse external library
$\mathcal{S}$, where each entry $S_q$ covers one empirically observed confusion pair $q$.
The initial library $\mathcal{S}^{(0)}$ is constructed before
evolution, while $g_{\mathrm{init}}$ is subsequently reused to
initialize new skills for recurrent uncovered pairs, as summarized
in Algorithm~\ref{alg:case}.
At evolution step $t$, CASE maintains a sparse external skill
library $\mathcal{S}^{(t)}$, whose domain
$\operatorname{dom}(\mathcal{S}^{(t)})$ contains only the confusion
pairs observed in the historical predictions of the frozen MLLM.
Rather than instantiating skills for all $\binom{K}{2}$ possible
emotion pairs, CASE constructs a pair-specific skill
$S_q^{(t)}$ only when the corresponding boundary $q$ has been
empirically observed. The resulting library therefore follows the
model's empirical emotion-confusion structure while remaining
sparse.

\begin{figure*}[t]
    \centering
    \IfFileExists{Figures/pipeline1.png}{%
        \includegraphics[width=\textwidth]{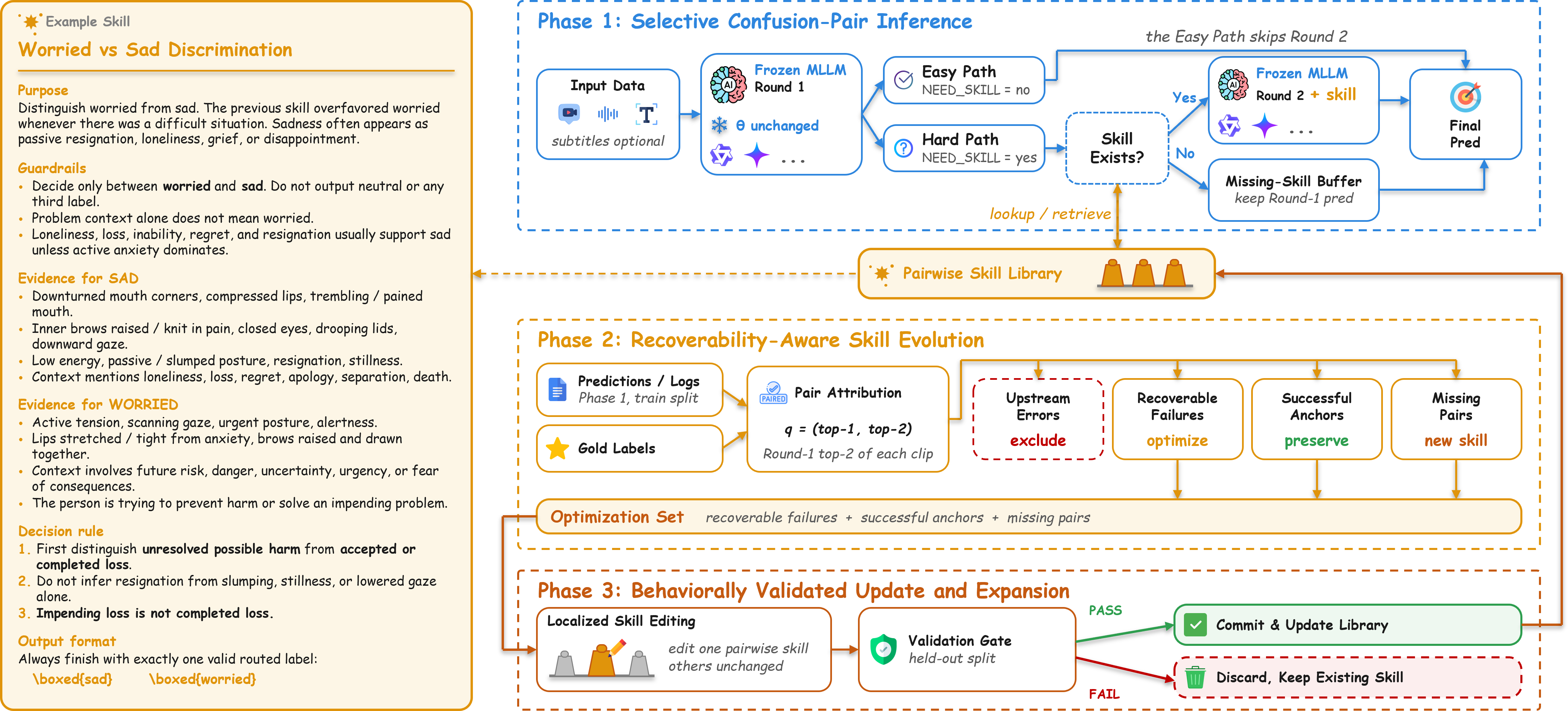}%
    }{%
        \fbox{%
            \parbox[c][0.22\textheight][c]{0.9\textwidth}{%
                \centering
                Method overview figure unavailable\\
                \texttt{Figures/pipeline1.png}
            }%
        }%
    }
    \caption{
        Overview of the CASE localize--attribute--validate
        lifecycle. A frozen MLLM selectively reopens one
        ambiguous emotion-pair boundary. After ground-truth
        feedback becomes available, candidate-recall failures
        are excluded from pair-level learning, while attributable
        failures and successful anchors guide a local EmotionSkill
        revision. The revision affects future emotion predictions
        only after pair-specific behavioral validation.
    }
    \label{fig:method}
\end{figure*}

As shown in Figure~\ref{fig:method}, CASE follows a
\textit{localize--attribute--validate} lifecycle: it selectively reopens an
ambiguous emotion boundary, attributes post-inference feedback to
that boundary, and deploys a skill revision only after
pair-specific behavioral validation. Algorithm~\ref{alg:case}
summarizes the sequential evolution procedure. Only the external
EmotionSkill library is updated; the MLLM remains frozen.

\begingroup
\definecolor{casegray}{gray}{0.46}
\algrenewcommand{\algorithmiccomment}[1]{%
  \hfill\textcolor{casegray}{$\triangleright$\,\emph{#1}}}
\begin{algorithm}[t]
\caption{CASE: Confusion-Aware Skill Evolution}
\label{alg:case}
\footnotesize
\begin{algorithmic}[1]
\Require Frozen MLLM $f_\theta$, editor $g_\phi$, initializer
$g_{\mathrm{init}}$, the skill library $S^{}$, evolution stream
$\mathcal{D}$, fixed local validation sets $\{\mathcal{V}_q\}$
\Ensure Evolved skill library $\mathcal{S}$
\State $\mathcal S \leftarrow \mathcal S^{(0)}$
\State Initialize empty failure, anchor, and missing-pair buffers
\ForAll{$(x,y^*)\in\mathcal{D}$}
    \State $(\hat y_1^{(1)},\hat y_2^{(1)},r,z)
    \gets f_\theta(x,p_{\mathrm{diag}})$;
    $q\gets\{\hat y_1^{(1)},\hat y_2^{(1)}\}$
    \If{$r=0$}
        \State \textbf{continue} \Comment{clear sample}
    \EndIf
    \State Run pairwise inference with $S_q$ if
    $q\in\operatorname{dom}(\mathcal{S})$
    \State After inference, reveal $y^*$ for trajectory attribution
    \If{$y^*\notin q$}
        \State \textbf{continue} \Comment{upstream candidate-recall error}
    \EndIf
    \If{$q\notin\operatorname{dom}(\mathcal{S})$}
        \State Add the diagnostic record to $\mathcal{M}_q$
        \State \Call{BootstrapExpand}{$q,\mathcal{M}_q,g_{\mathrm{init}},
        \mathcal{V}_q,\mathcal{S}$} \Comment{secondary}
        \State \textbf{continue}
    \EndIf
    \Statex \hspace{\algorithmicindent}
    Primary: recoverability-guided local evolution
    \State Add the trajectory to $\mathcal{B}_q^+$ if
    $\hat y^{(2)}=y^*$, otherwise to $\mathcal{B}_q^-$
    \If{$\mathcal{B}_q^-$ is not ready for evolution}
        \State \textbf{continue}
    \EndIf
   \State Select bounded sets
$F_q\subseteq B_q^-$ and $P_q\subseteq B_q^+$
\State $\Delta_q
\leftarrow g_\phi(S_q,F_q,P_q)$
\State $\widetilde S_q
\leftarrow\operatorname{Apply}(S_q,\Delta_q)$
    \If{$\widetilde S_q$ strictly improves on $\mathcal{V}_q$
    without prediction collapse}
        \State $S_q\gets\widetilde S_q$ \Comment{commit the local patch}
    \EndIf
\State $B_q^-\leftarrow B_q^-\setminus F_q$
\Comment{consume the edited failures}
\EndFor
\State \Return $\mathcal{S}$
\end{algorithmic}
\end{algorithm}
\endgroup

\subsection{Phase 1: Selective Confusion-Pair Inference}
\label{sec:selective_inference}

Applying pair-specific EmotionSkills indiscriminately may disturb
emotion predictions that are already supported by clear
multimodal evidence. CASE therefore intervenes only when the
frozen MLLM identifies a specific unresolved emotion boundary.
This selective design preserves confident predictions while
concentrating additional reasoning on ambiguous samples.

For each input sample $x$, the MLLM receives a diagnostic
prompt $p_{\mathrm{diag}}$ and produces
\begin{equation}
\left(
\hat y_1^{(1)},
\hat y_2^{(1)},
r,
z
\right)
=
f_\theta
\left(
x,
p_{\mathrm{diag}}
\right).
\label{eq:diagnostic}
\end{equation}
The corresponding candidate confusion pair is
\[
q=q(x)=
\left\{\hat y_1^{(1)},\hat y_2^{(1)}\right\},
\]
where $\hat y_1^{(1)}$ and $\hat y_2^{(1)}$ are two distinct
emotion candidates in round-1, $r\in\{0,1\}$ is the routing decision, and
$z$ records the diagnostic multimodal evidence.

The model returns $r=1$ when multimodal evidence does not reliably distinguish the two candidate
emotions. If $r=0$, CASE treats the
emotion as sufficiently clear and directly preserves the
first-round prediction $\hat y_1^{(1)}$.

If $r=1$ and the corresponding skill $S_q^{(t)}$ exists, CASE
reopens only this emotion boundary through a second multimodal
inference:
\begin{equation}
\hat y^{(2)}
=
f_\theta
\left(
x,
p_{\mathrm{pair}}(q)
\oplus
S_q^{(t)}
\right),
\qquad
\hat y^{(2)}\in q,
\label{eq:pairwise_inference}
\end{equation}
where $\oplus$ denotes prompt concatenation. The model receives
the original $x$ again, while
$p_{\mathrm{pair}}(q)$ restricts the output space to the two
candidate emotions.

The EmotionSkill provides a pair-specific discrimination protocol that guides the MLLM to re-examine relevant facial, behavioral, temporal, and
contextual cues, compare the evidence supporting both emotions,
and check misleading or shared cues before making the final
choice. This
turns the original $K$-class prediction into a focused
emotion-boundary comparison grounded in the input sample.

Accordingly, the three inference outcomes match the three paths in
Phase~1 of Figure~\ref{fig:method}: clear samples retain
$\hat y_1^{(1)}$, routed samples with a covered pair use
$\hat y^{(2)}$, and routed samples with an uncovered pair fall back
to $\hat y_1^{(1)}$ while logging the missing boundary.
The covered-pair path yields a round-2 trajectory for skill evolution in Phase 2, whereas the uncovered-pair path retains the diagnostic record as evidence for sparse library expansion.
\subsection{Phase 2: Recoverability-Aware Skill Evolution}

\paragraph{Trajectory attribution.}
For each routed sample with a covered pair, CASE records a trajectory $\tau$ comprising the diagnostic evidence, active skill version, first- and second-round
predictions, and ground-truth feedback. We determine whether the
result can be attributed to the active pair using
\begin{equation}
\rho(\tau)=\mathbb{I}[y^*\in q].
\end{equation}
Only trajectories with $\rho(\tau)=1$ provide pair-level learning
signals. Incorrect second-round predictions are stored as recoverable
failures in $B_q^-$, while correct predictions are retained in
$B_q^+$ as successful anchors. Trajectories with $\rho(\tau)=0$ are
candidate-recall errors and are excluded from EmotionSkill editing.

\paragraph{Contrastive textual editing.}
When sufficient failures have accumulated, CASE selects a bounded
failure set $F_q\subseteq B_q^-$ and anchor set
$P_q\subseteq B_q^+$. The editor proposes a local textual patch:
\begin{equation}
\Delta_q=g_\phi(S_q,F_q,P_q),\qquad
\widetilde S_q=\operatorname{Apply}(S_q,\Delta_q).
\end{equation}
The patch extracts reusable distinctions, preserves valid rules
supported by the anchors, and avoids sample-specific or
label-forcing expressions.
\subsection{Phase 3: Behaviorally Validated Update and Expansion}

\noindent\textbf{Validated skill update.}
A candidate EmotionSkill revision is not activated immediately, since a
textually plausible rule may still overcorrect the targeted failures or
bias the pairwise prediction toward one emotion. CASE therefore
evaluates the current and candidate EmotionSkills on the same fixed
pair-specific validation subset. The candidate
is committed only when it strictly improves the pair-specific
predictions without collapsing toward either label; otherwise, the
existing EmotionSkills are retained. This behavioral gate ensures that only
revisions with measurable corrective effects enter the active library.

\noindent\textbf{Sparse skill expansion.}
The missing-pair path in Figure~\ref{fig:method} handles
routed boundaries that are not yet covered by the active library.
As summarized by \Call{BootstrapExpand}{} in
Algorithm~\ref{alg:case}, an uncovered pair contributes expansion evidence
only when the ground-truth emotion belongs to that pair.
Once sufficient
recurrent evidence has accumulated, the initializer proposes a new
pair-specific EmotionSkill. Because no active parent EmotionSkill exists for direct
comparison, the candidate is evaluated by a standalone bootstrap gate
and is activated only when it has sufficient validation support and
does not produce label collapse. Rejected candidates remain inactive.
Consequently, the library expands only toward recurrent and
behaviorally supported confusion boundaries rather than exhaustively
enumerating all possible emotion pairs.

Detailed validation-set construction, behavioral metrics, acceptance
criteria, and implementation thresholds are provided in the
supplementary Sec. A.


\section{Experiments}
\label{sec:experiments}
We evaluate CASE through four research questions:

\noindent\textbf{(RQ1)}
Does the complete CASE pipeline remain effective across different backbones and tasks?\\
\noindent\textbf{(RQ2)}
Where does selective routing concentrate its interventions?\\
\noindent\textbf{(RQ3)}
Does trajectory-driven evolution improve a static pairwise EmotionSkill library?\\
\noindent\textbf{(RQ4)}
Does behaviorally validated deployment filter unsupported revisions and
expand skill coverage without exhaustive library growth?

\subsection{Experimental Setup}
\label{sec:experimental_setup}

\paragraph{Datasets and metrics.}
 We evaluate CASE across three complementary settings: cross-modal emotion conflicts on CA-MER~\citep{han2025camer}, in-the-wild dynamic facial expression recognition on DFEW~\citep{jiang2020dfew}, and fine-grained multimodal emotion reasoning on EMER~\citep{lian2023emer}. CA-MER contains the Audio-Conflict, Visual-Conflict, and Consistent subsets, for which we report accuracy (Acc.). For DFEW, we report unweighted average recall (UAR) and weighted average recall (WAR), while accuracy is used for EMER.

\paragraph{Backbones and implementation.}
We evaluate CASE with three frozen backbones: Qwen3-VL-8B~\citep{bai2025qwen3vl}, Emotion-Qwen~\citep{huang2025emotionqwen}, and Gemini-3.5-Flash~\citep{googledeepmind2026gemini35flash}. GPT-4o~\citep{openai2024gpt4o} or GPT-5.6-sol~\citep{openai2026gpt56sol} serves only as the external skill editor. CASE-Full injects the complete routed skill, whereas CASE-Retrieval retrieves only relevant skill fragments. Evolution, validation, and test sets are strictly disjoint. Further implementation details are provided in Supplementary Sec. B.

\subsection{Overall Effectiveness and Generality (RQ1)}

We first examine the in-distribution
effectiveness of CASE on DFEW, where the skill library is
constructed. We then evaluate CASE on CA-MER, a benchmark specifically designed to assess emotion recognition under conflicting multimodal cues. Finally,
we evaluate CASE on the open-set EMER benchmark, which
requires fine-grained emotion reasoning beyond a fixed
closed-set label space. 

\begin{table*}[t]
    \centering

    \renewcommand{\arraystretch}{1.12}
    \setlength{\tabcolsep}{4.5pt}
    \small

    \begin{adjustbox}{max width=\textwidth}
    \begin{tabular}{lllcccccc}
        \toprule

        \multirow{2}{*}{\textbf{Inference Backbone}}
        & \multirow{2}{*}{\textbf{Evolution LLM}}
        & \multirow{2}{*}{\textbf{Variant}}
        & \multicolumn{4}{c}{\textbf{CA-MER}}
        & \multicolumn{2}{c}{\textbf{DFEW}} \\

        \cmidrule(lr){4-7}
        \cmidrule(lr){8-9}

        &
        &
        & \textbf{Audio-Conflict}
        & \textbf{Visual-Conflict}
        & \textbf{Consistent}
        & \textbf{Overall}
        & \textbf{UAR}
        & \textbf{WAR} \\

        \midrule

        \multirow{4}{*}{Qwen3-VL-8B}
        & --
        & Frozen backbone
        & 31.00
        & 48.80
        & 60.60
        & 46.80
        & 51.59
        & 59.42 \\

        & \multirow{2}{*}{GPT-4o}
        & CASE-Full
        & \textbf{49.00}\gain{18.00}
        & \textbf{55.40}\gain{6.60}
        & \textbf{73.00}\gain{12.40}
        & \textbf{59.13}\gain{12.33}
        & 49.68\loss{1.91}
        & 58.08\loss{1.34} \\

        &
        & CASE-Retrieval
        & 47.00\gain{16.00}
        & \textbf{55.40}\gain{6.60}
        & 72.80\gain{12.20}
        & 58.40\gain{11.60}
        & \textbf{55.30}\gain{3.71}
        & \textbf{61.21}\gain{1.79} \\

        & GPT-5.6-sol
        & CASE-Full
        & \textbf{49.00}\gain{18.00}
        & 54.40\gain{5.60}
        & 72.00\gain{11.40}
        & 58.47\gain{11.67}
        & 53.02\gain{1.43}
        & 60.49\gain{1.07} \\

        \midrule

        \multirow{3}{*}{Emotion-Qwen}
        & --
        & Frozen backbone
        & 47.70
        & 59.72
        & \textbf{72.95}
        & 60.12
        & 56.64
        & 66.17 \\

        & \multirow{2}{*}{GPT-4o}
        & CASE-Full
        & 48.97\gain{1.27}
        & 60.45\gain{0.73}
        & 65.63\loss{7.32}
        & 58.35\loss{1.77}
        & \textbf{57.60}\gain{0.96}
        & \textbf{67.36}\gain{1.19} \\

        &
        & CASE-Retrieval
        & \textbf{49.27}\gain{1.57}
        & \textbf{62.13}\gain{2.41}
        & 72.48\loss{0.47}
        & \textbf{61.30}\gain{1.18}
        & 56.74\gain{0.10}
        & 65.01\loss{1.16} \\

        \midrule

        \multirow{3}{*}{Gemini-3.5-Flash}
        & --
        & Frozen backbone
        & 56.40
        & 63.60
        & 80.20
        & 66.73
        & 54.49
        & 60.92 \\

        & \multirow{2}{*}{GPT-5.6-sol}
        & CASE-Full
        & \textbf{62.00}\gain{5.60}
        & \textbf{64.40}\gain{0.80}
        & \textbf{85.40}\gain{5.20}
        & \textbf{70.60}\gain{3.87}
        & \textbf{60.88}\gain{6.39}
        & \textbf{68.69}\gain{7.77} \\

        &
        & CASE-Retrieval
        & 59.60\gain{3.20}
        & 61.80\loss{1.80}
        & 81.60\gain{1.40}
        & 67.67\gain{0.94}
        & 58.91\gain{4.42}
        & 67.79\gain{6.87} \\

        \bottomrule
    \end{tabular}
    \end{adjustbox}
     \caption{
        Backbone-wise effectiveness of CASE across the CA-MER
        and DFEW benchmarks.
    }
    \label{tab:backbone_effectiveness}
\end{table*}

\begin{table}[t]
    \centering

    \renewcommand{\arraystretch}{1.10}
    \setlength{\tabcolsep}{5pt}
    \small

    \begin{adjustbox}{max width=\columnwidth}
    \begin{tabular}{lcc}
        \toprule
        \textbf{Model}
        & \textbf{UAR}
        & \textbf{WAR} \\
        \midrule

        MoSEAR~\cite{han2025camer}
        & 32.94
        & 42.67 \\

        VideoAuto-R1~\cite{liu2026videoautor1}
        & 55.78
        & 64.45 \\

        Emotion-LLaMA~\cite{cheng2024emotionllama}
        & 45.59
        & 59.37 \\

        Video-LLaMA~\cite{zhang2023videollama}
        & 26.77
        & 35.75 \\

        Video-LLaVA (video)~\cite{lin2024videollava}
        & 26.17
        & 35.24 \\

        Qwen-Audio~\cite{chu2023qwenaudio}
        & 25.23
        & 31.74 \\

        Qwen2.5-VL-7B~\cite{bai2025qwen25vl}
        & 43.28
        & 50.36 \\

        Qwen3-VL-8B~\cite{bai2025qwen3vl}
        & 51.59
        & 59.42 \\

        Emotion-Qwen~\cite{huang2025emotionqwen}
        & 56.64
        & 66.17 \\

        Gemini-3.5-Flash~\cite{googledeepmind2026gemini35flash}
        & 54.49
        & 60.92 \\

        \midrule
        \rowcolor{gray!15}
        \multicolumn{3}{l}{\textit{Ours}} \\

        CASE (Qwen3-VL-8B)$^{\dagger}$
        & 55.30
        & 61.21 \\

        CASE (Emotion-Qwen)$^{\dagger}$
        & 57.60
        & 67.36 \\

        CASE (Gemini-3.5-Flash)$^{\ddagger}$
        & \textbf{60.88}
        & \textbf{68.69} \\

        \bottomrule
    \end{tabular}
    \end{adjustbox}
      \caption{
        Performance comparison on the DFEW benchmark.
        For each inference backbone and evaluation metric,
        we report the better result between CASE-Full and
        CASE-Retrieval.
        $\dagger$ denotes skill evolution with GPT-4o, and
        $\ddagger$ denotes skill evolution with GPT-5.6-sol.
    }
    \label{tab:dfew_comparison}
\end{table}
\begin{figure}[t]
    \centering
    \includegraphics[width=\columnwidth]{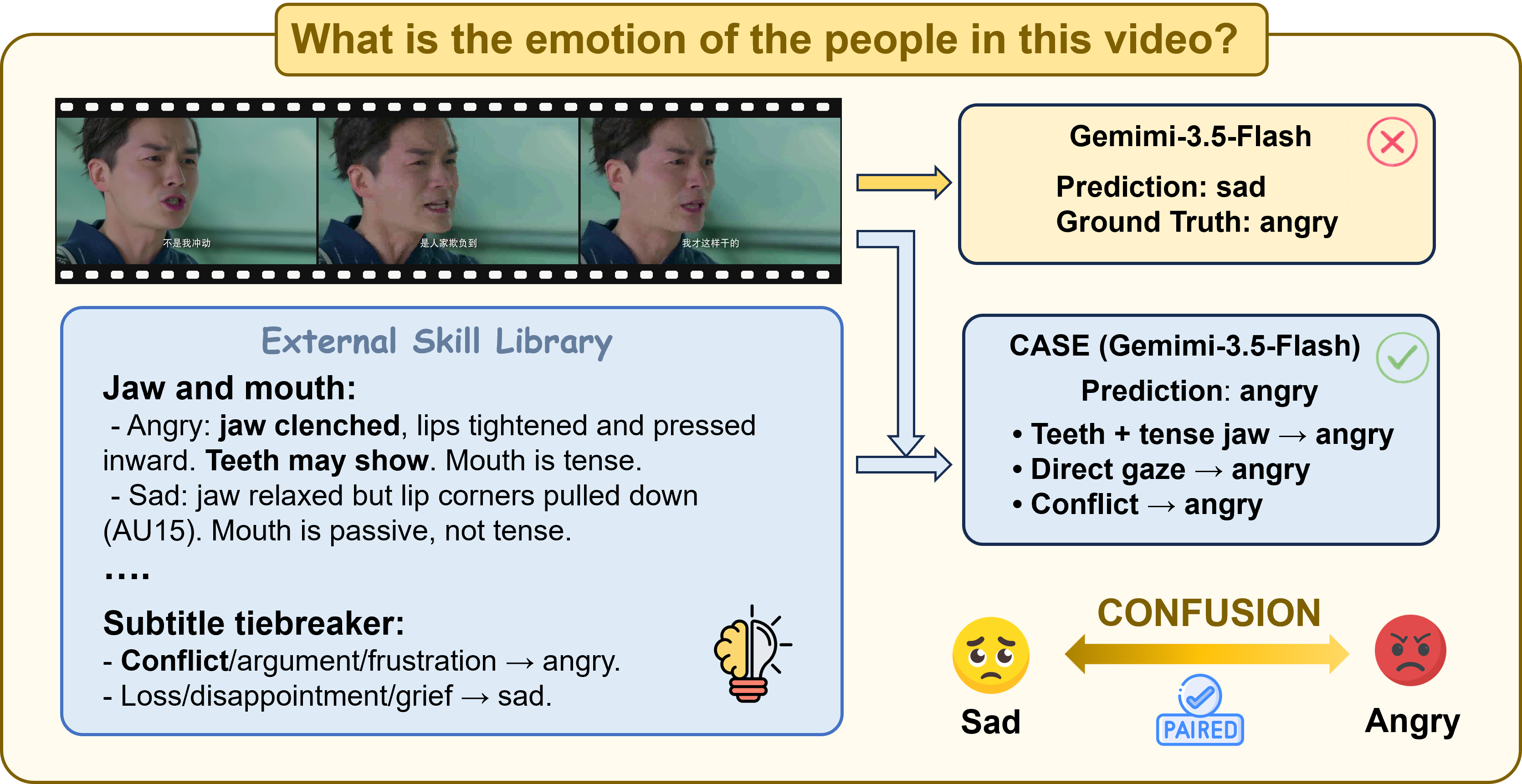}
    \caption{
An example of EmotionSkill-guided pairwise repair in CASE.
The backbone confuses \emph{angry} with \emph{sad}, guided by tension-related facial cues and conflict context, the second round corrects the prediction to
\emph{angry}.
}
    \label{fig:qualitative_case}
\end{figure}
\paragraph{Qualitative example of pairwise repair.}
Figure~\ref{fig:qualitative_case} shows how CASE corrects a
\emph{sad}--\emph{angry} confusion. The backbone initially predicts
\emph{sad}, whereas the pair-specific EmotionSkill guides a focused
re-evaluation and corrects the prediction to \emph{angry}. This
demonstrates how CASE repairs an ambiguous local boundary without
reconsidering the full label space.
\paragraph{The improvements on DFEW.}
As shown in Tables~\ref{tab:backbone_effectiveness}
and~\ref{tab:dfew_comparison}, CASE consistently benefits
different frozen backbones. In particular, CASE with
Gemini-3.5-Flash achieves 60.88 UAR and 68.69 WAR,
establishing the effectiveness of pair-specific skills in the
in-distribution setting.

\paragraph{Cross-task generalization on CA-MER.}
\begin{table*}[t]
    \centering

    \renewcommand{\arraystretch}{1.10}
    \setlength{\tabcolsep}{8pt}
    \small

    \begin{adjustbox}{max width=\textwidth}
    \begin{tabular}{lcccc}
        \toprule

        \textbf{Model}
        & \textbf{Audio-Conflict}
        & \textbf{Visual-Conflict}
        & \textbf{Consistent}
        & \textbf{Overall} \\

        \midrule

        \rowcolor{gray!15}
        \multicolumn{5}{l}{\textbf{Closed-source Models}} \\

        Gemini-3.5-Flash~\cite{googledeepmind2026gemini35flash}
        & 56.40
        & \underline{63.60}
        & \underline{80.20}
        & \underline{66.73} \\

        \rowcolor{gray!15}
        \multicolumn{5}{l}{\textbf{Open-source Models}} \\

        mPLUG-Owl3~\cite{ye2025mplugowl3}
        & 41.08
        & 40.71
        & 49.29
        & 43.69 \\

        Chat-UniVi~\cite{jin2024chatunivi}
        & 42.28
        & 43.43
        & 50.38
        & 45.36 \\

        VideoAuto-R1~\cite{liu2026videoautor1}
        & 43.60
        & 47.40
        & 67.60
        & 52.87 \\

        Emotion-LLaMA~\cite{cheng2024emotionllama}
        & \underline{59.94}
        & 47.66
        & 59.23
        & 55.61 \\

        Qwen2.5-VL-7B~\cite{bai2025qwen25vl}
        & 37.27
        & 44.20
        & 63.93
        & 48.47 \\

        Qwen3-VL-8B~\cite{bai2025qwen3vl}
        & 31.00
        & 48.80
        & 60.60
        & 46.80 \\

        Emotion-Qwen~\cite{huang2025emotionqwen}
        & 47.70
        & 59.72
        & 72.95
        & 60.12 \\

        \rowcolor{gray!15}
        \multicolumn{5}{l}{\textbf{Ours}} \\

        CASE (Qwen3-VL-8B)$^{\dagger}$
        & 49.00
        & 55.40
        & 73.00
        & 59.13 \\

        CASE (Emotion-Qwen)$^{\dagger}$
        & 48.97
        & 60.45
        & 65.63
        & 58.35 \\

        CASE (Gemini-3.5-Flash)$^{\ddagger}$
        & \textbf{62.00}
        & \textbf{64.40}
        & \textbf{85.40}
        & \textbf{70.60} \\

        \bottomrule
    \end{tabular}
    \end{adjustbox}
     \caption{
        Comparison of CASE with existing methods on the CA-MER dataset.
        For each inference backbone, we report the CASE-Full configuration.
        The skill-retrieval variants are analyzed separately in
        Table~\ref{tab:backbone_effectiveness}.
    }
    \label{tab:camer_comparison}
\end{table*}
CA-MER's conflicting multimodal cues naturally induce local
emotion-boundary ambiguity, making it well suited to CASE.
As shown in Tables~\ref{tab:backbone_effectiveness} and
\ref{tab:camer_comparison}, CASE-Full improves Qwen3-VL-8B
from 46.80\% to 59.13\% and Gemini-3.5-Flash from 66.73\%
to 70.60\%, while Emotion-Qwen benefits most from
CASE-Retrieval. These results demonstrate the generalization of CASE across backbones and tasks. Complete accuracy, macro-recall, and averaged results for all
backbones and CA-MER subsets are reported in Supplementary Sec. C.
\begin{figure}[t]
    \centering
    \includegraphics[width=\columnwidth]{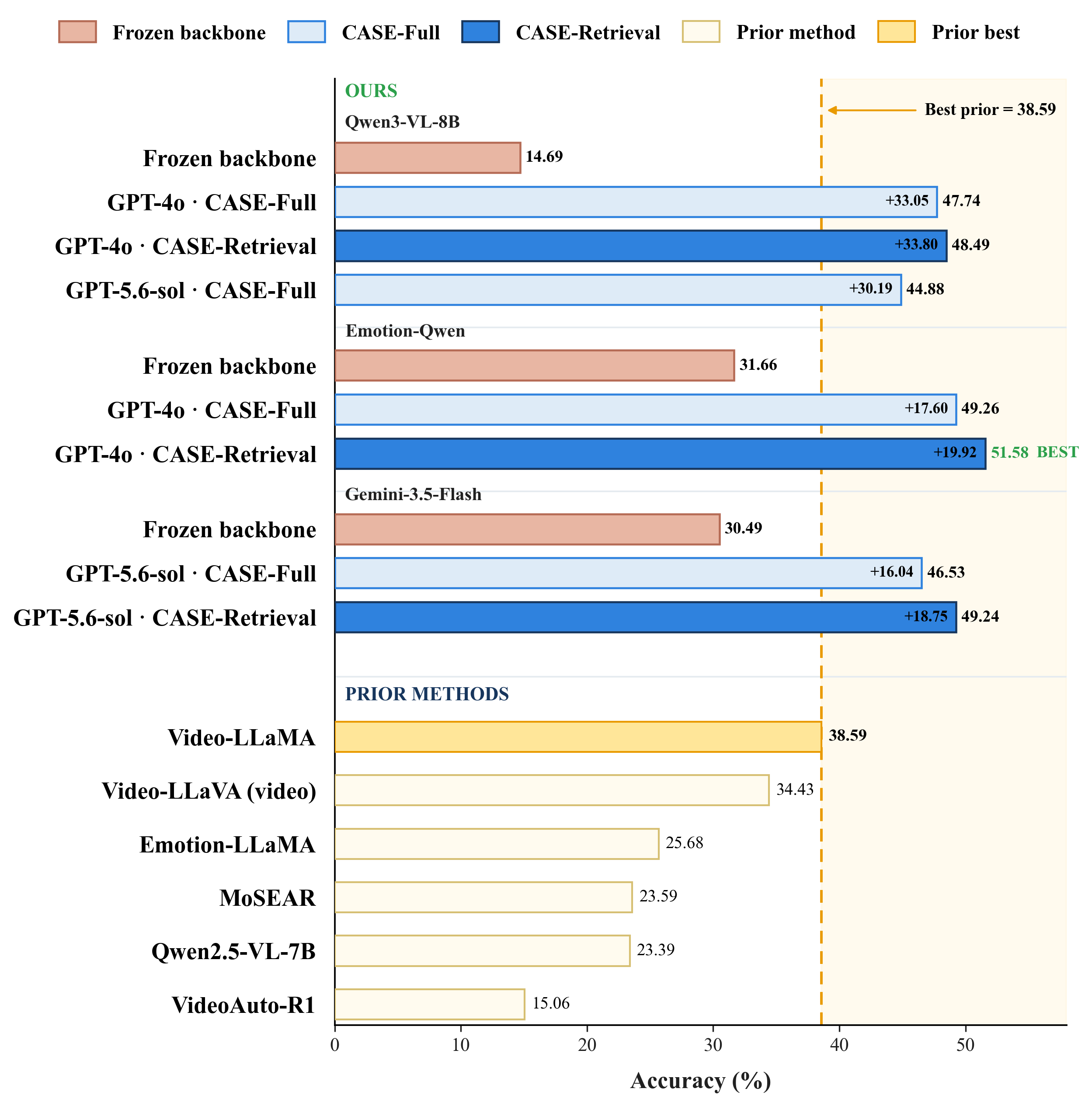}
    \caption{
        Open-set generalization on EMER. CASE improves all three
        frozen backbones without EMER-specific adaptation, with
        the best result achieved by CASE-Retrieval with
        Emotion-Qwen.
    }
    \label{fig:emer_open_set}
\end{figure}
\paragraph{Open-set generalization on EMER.}
On open-vocabulary EMER (Figure~\ref{fig:emer_open_set}), CASE improves all three frozen backbones, with CASE-Retrieval consistently outperforming CASE-Full. Emotion-Qwen with CASE-Retrieval achieves 51.58\%, surpassing prior methods and demonstrating effective cross-backbone transfer. Moreover, substantial gains over the
backbones are observed for all three inference models,
demonstrating that CASE generalizes effectively across
backbones and open-set emotion reasoning settings.
\begin{table*}[t]
    \centering
   
    \renewcommand{\arraystretch}{1.15}
    \setlength{\tabcolsep}{6pt}
    \small
    \begin{adjustbox}{max width=\textwidth}
    \begin{tabular}{lcccccccc}
        \toprule

        \multirow{2}{*}{\textbf{Variant}}
        & \multicolumn{4}{c}{\textbf{Components}}
        & \multicolumn{4}{c}{\textbf{CA-MER}} \\

        \cmidrule(lr){2-5}
        \cmidrule(lr){6-9}

        & \textbf{Sel.}
        & \textbf{skill}
        & \textbf{Evo.}
        & \textbf{Gate}
        & \textbf{Audio-Conflict}
        & \textbf{Visual-Conflict}
        & \textbf{Consistent}
        & \textbf{Overall} \\

        \midrule

        Vanilla MLLM
        & -- & -- & -- & --
        & 31.00
        & 48.80
        & 60.60
        & 46.80 \\

        + Selective Reasoning
        & \cmark & -- & -- & --
        & 32.60
        & 48.40
        & 61.40
        & 47.47 \\

        + Static Pairwise EmotionSkills
        & \cmark & \cmark & -- & --
        & 45.80
        & 54.00
        & 71.20
        & 57.00 \\

        + Ungated EmotionSkill Evolution
        & \cmark & \cmark & \cmark & --
        & 47.80
        & 54.00
        & 73.00
        & 58.27 \\

        \rowcolor{gray!15}
        \textbf{CASE}
        & \cmark & \cmark & \cmark & \cmark
        & \textbf{49.00}
        & \textbf{55.40}
        & \textbf{73.00}
        & \textbf{59.13} \\

        \bottomrule
    \end{tabular}
    \end{adjustbox}
     \caption{
        Ablation study of CASE on the CA-MER dataset.
        \textbf{Sel.} denotes selective confusion-pair reasoning;
        \textbf{skill} denotes pairwise EmotionSkill injection;
        \textbf{Evo.} denotes recoverability-aware, trajectory-driven
        skill evolution;
        and \textbf{Gate} denotes behavioral validation and library
        expansion.
        \textbf{Overall} is the average accuracy across the three subsets.
    }
    \label{tab:case_ablation}

\end{table*}
\subsection{Selective Local Repair (RQ2)}
\label{sec:selective_localization}


\paragraph{Does routing identify useful intervention targets?}

\begin{figure}[t]
    \centering
    \includegraphics[width=\columnwidth]{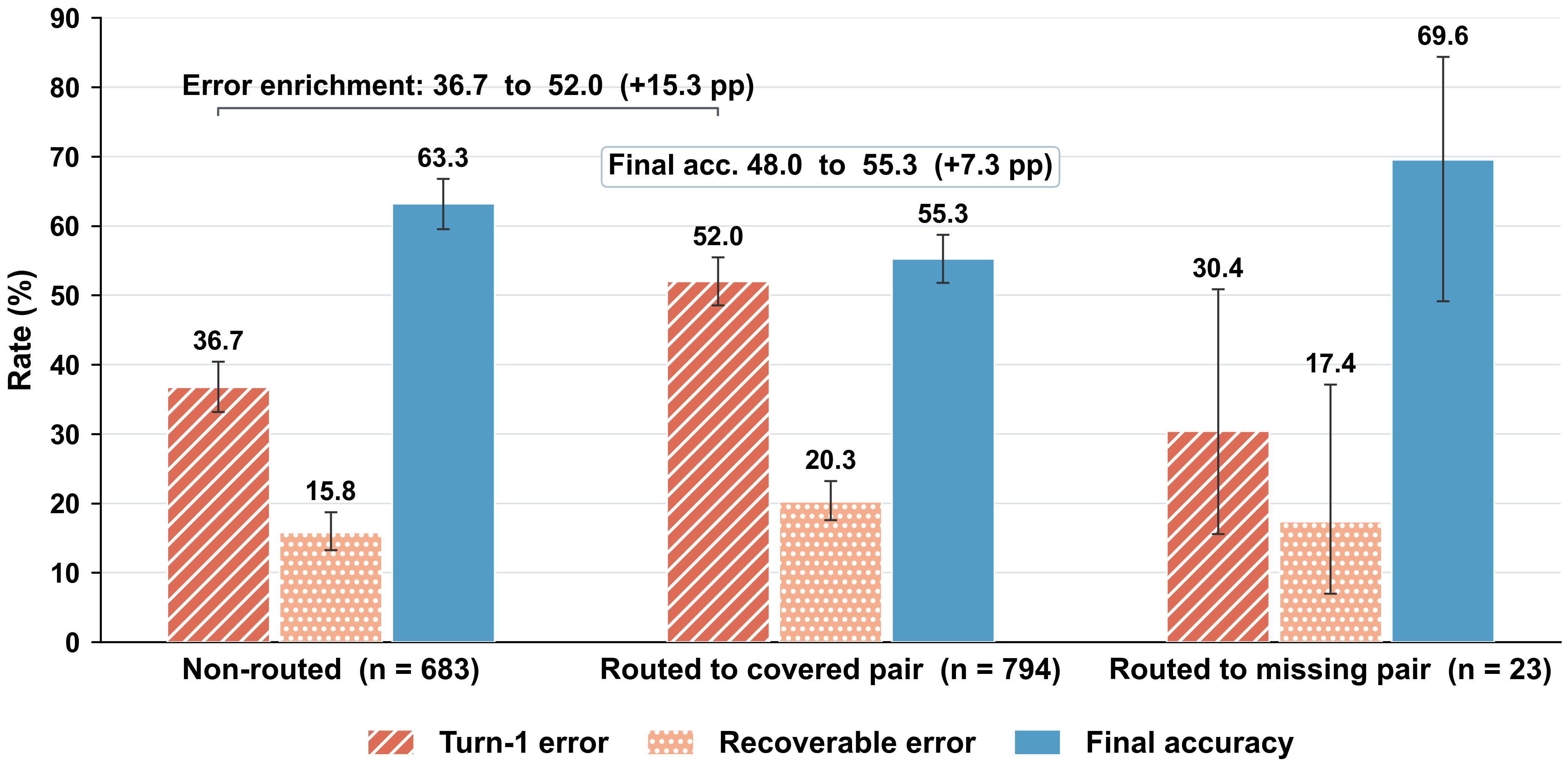}
    \caption{
        Effectiveness across routing groups on CA-MER.
        Test samples are partitioned according to the routing
        decision and whether the confusion pair is covered by the
        skill library.
    }
    \label{fig:routing_group_effectiveness}
\end{figure}

Figure~\ref{fig:routing_group_effectiveness} shows that routing
concentrates on more difficult samples. The routed-to-covered-pair
group has a round-1 error rate of 52.0\%, compared with 36.7\%
for non-routed samples. After pairwise intervention, its accuracy
increases from 48.0\% to 55.3\%, a gain of 7.3 percentage points. In addition, detailed Repair Rate, Harm Rate, and Repair--Harm Gap results
are reported in Supplementary Sec. D, where all configurations
achieve positive overall gaps.

Non-routed samples retain their round-1 predictions and an accuracy
of 63.3\%. Samples routed to uncovered pairs are likewise preserved,
with a round-1 accuracy of 69.6\%. CASE therefore intervenes mainly
on difficult, covered cases while avoiding unnecessary changes
elsewhere.

\begin{figure}[t]
    \centering
    \includegraphics[
        width=\columnwidth
    ]{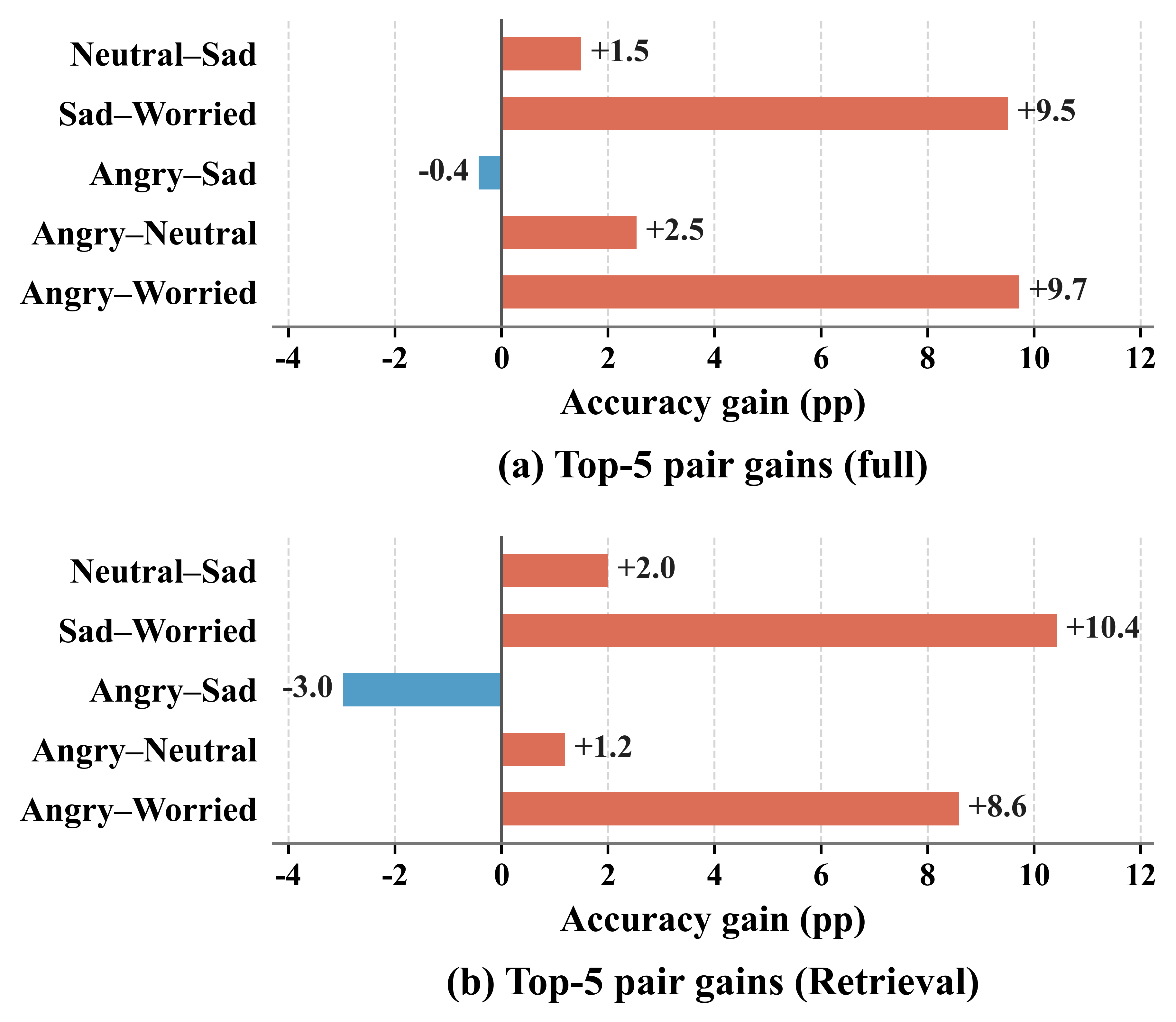}
    \caption{
         PairAcc changes on the five most frequent first-round
        confusion pairs under CASE-Full and CASE-Retrieval.
    }
    \label{fig:case_improvement_analysis}
\end{figure}

\paragraph{Which confusion boundaries are repaired?}

For a confusion pair $q=\{y_a,y_b\}$, let
\begin{equation}
\mathcal{D}_q
=
\left\{
i \mid y_i^* \in q
\right\}
\end{equation}
denote the samples whose ground-truth label belongs to $q$.
For $m\in\{\mathrm{R1},\mathrm{CASE}\}$, pair-specific accuracy is
\begin{equation}
\mathrm{PairAcc}^{(m)}(q)
=
\frac{
\sum_{i\in\mathcal{D}_q}
\mathbb{I}
\left[
\hat{y}_i^{(m)}=y_i^*
\right]
}{
|\mathcal{D}_q|
},
\label{eq:pairwise_analysis_accuracy}
\end{equation}
and the change introduced by CASE is
\begin{equation}
\Delta\mathrm{PairAcc}(q)
=
\mathrm{PairAcc}^{(\mathrm{CASE})}(q)
-
\mathrm{PairAcc}^{(\mathrm{R1})}(q).
\label{eq:pairwise_accuracy_gain}
\end{equation}

Figure~\ref{fig:case_improvement_analysis} shows that both CASE variants improve most of the confusion pairs, especially \textit{sad--worried} and \textit{angry--worried}. Further analyses are provided in Supplementary. Complete class-transition statistics, subset-wise recall gains Supplementary
Sec. C.
\subsection{Trajectory-Driven Skill Evolution (RQ3)}
\label{sec:skill_evolution}

\paragraph{Evolution beyond static skills.}
Table~\ref{tab:case_ablation} shows that trajectory-driven evolution improves the
static EmotionSkill library from 57.00\% to 58.27\% overall accuracy.
More behaviorally validated skill evolution visualization can be found in Supplementary Sec. E. 
Together, these results demonstrate
that the external library can improve from prediction
trajectories while the backbone remains frozen.
 

\subsection{Validated Deployment and Sparse Expansion (RQ4)}
\label{sec:validated_deployment} 
Among 114 candidate revisions, CASE accepts 10, rejecting 102 without validation gains and two causing label collapse. This low acceptance rate is intentional: the editor acts as a proposal generator rather than an authority that directly modifies the deployed system. The validation gate separates textual plausibility from behavioral utility and prevents unsupported or label-collapsing revisions from entering the active library. Thus, the purpose of the gate is not to maximize the number of accepted revisions, but to admit only behaviorally supported updates. Four new pairs pass the bootstrap gate, expanding coverage from 17 to 21 pairs and improving CA-MER accuracy from 58.27\% to 59.13\% in Table \ref{tab:case_ablation}. This demonstrates controlled skill updating and sparse expansion. Detailed acceptance dynamics,
pair-specific validation changes, and expansion events are shown
in Supplementary Sec. E.

\section{Conclusion}
In this paper, we introduced CASE, a confusion-aware framework for improving frozen MLLMs in multimodal emotion boundary discrimination. Rather than globally updating model parameters, CASE localizes ambiguous predictions to recurrent emotion-pair boundaries and selectively invokes pair-specific EmotionSkills for focused discrimination.
Experiments across three emotion tasks and multiple backbones demonstrate consistent effectiveness and generality. These results highlight external skill evolution as a promising alternative to parameter-based adaptation for multimodal emotion understanding.

\clearpage
\clearpage
\bibliography{reference}
\end{document}